\pdfoutput=1

\documentclass[11pt]{article}

\usepackage{ACL2023}

\usepackage{times}
\usepackage{latexsym}
\usepackage{booktabs}
\usepackage{graphicx}

\usepackage[T1]{fontenc}

\usepackage[utf8]{inputenc}

\usepackage{microtype}

\usepackage{inconsolata}

\title{Duluth at SemEval-2026 Task 6: DeBERTa with LLM-Augmented Data for Unmasking Political Question Evasions}

\author{
    \textbf{Shujauddin Syed} \& \textbf{Ted Pedersen} \\
    Department of Computer Science \\
    University of Minnesota \\
    Duluth, MN 55812 USA \\
    \texttt{\{syed0093, tpederse\}@d.umn.edu}
}

\begin{document}
\maketitle

\begin{abstract}
This paper presents the Duluth approach to SemEval-2026 Task 6 on CLARITY: Unmasking Political Question Evasions. We address Task 1 (clarity-level classification) and Task 2 (evasion-level classification), both of which involve classifying question--answer pairs from U.S.\ presidential interviews using a two-level taxonomy of response clarity. Our system is based on DeBERTa-V3-base, extended with focal loss, layer-wise learning rate decay, and boolean discourse features. To address class imbalance in the training data, we augment minority classes using synthetic examples generated by Gemini 3 and Claude Sonnet 4.5. Our best configuration achieved a Macro F1 of 0.76 on the Task 1 evaluation set, placing 8th out of 40 teams. The top-ranked system (TeleAI) achieved 0.89, while the mean score across participants was 0.70. Error analysis reveals that the dominant source of misclassification is confusion between Ambivalent and Clear Reply responses, a pattern that mirrors disagreements among human annotators. Our findings demonstrate that LLM-based data augmentation can meaningfully improve minority-class recall on nuanced political discourse tasks.
\end{abstract}

\section{Introduction}

The SemEval 2026 Task 6 on CLARITY: Unmasking Political Question Evasions, challenges participants to automatically detect and classify evasive communication strategies in political discourse.


The task is organized around a two-level taxonomy of response clarity \citep{thomas2024isaidthatdataset}: Task 1 requires classifying question--answer pairs from U.S.\ presidential interviews into three clarity categories (Clear Reply, Ambivalent, Clear Non-Reply), while Task 2 targets nine fine-grained evasion techniques.

Our system\footnote{Our code is publicly available at \url{https://github.com/syed0093-umn/SemEval2026_Task6_Duluth}.} is built on DeBERTa-V3-base \citep{he2023debertav}, extended with focal loss \citep{lin2017focal}, layer-wise learning rate decay, and two boolean discourse features extracted from the original data. To address severe class imbalance in the training set (59\% Ambivalent, 31\% Clear Reply, 10\% Clear Non-Reply), we augmented the minority classes using synthetic examples generated by Gemini 3 \footnote{\url{https://storage.googleapis.com/deepmind-media/Model-Cards/Gemini-3-Pro-Model-Card.pdf}} and Claude Sonnet 4.5\footnote{\url{https://www-cdn.anthropic.com/963373e433e489a87a10c823c52a0a013e9172dd.pdf}}.

Our Gemini-augmented system achieved a Macro F1 of 0.76 on the evaluation set for Task 1, placing 8th out of 40 teams. Error analysis reveals that the primary challenge lies in distinguishing Ambivalent responses from Clear Replies, a confusion that mirrors disagreements observed among human annotators.

\section{Task Description}

This task addresses the computational detection and classification of evasive communication strategies in political discourse. The task is grounded in well-established theories of equivocation from political science \citep{bullequi, bullqa, bull2019cant}, which show that politicians provide clear responses to only 39--46\% of questions during televised interviews. The shared task comprises two subtasks applied to question--answer (QA) pairs extracted from U.S.\ presidential interviews, organized around a two-level hierarchical taxonomy of response clarity proposed by \citet{thomas2024isaidthatdataset}.

\paragraph{Task 1 -- Clarity-level Classification:}

Given a question--answer pair from a political interview, classify the response into one of three clarity categories:

\textbf{Clear Reply}: \textit{The requested information is explicitly provided.}

\textbf{Ambivalent}: \textit{A response is given but allows for multiple interpretations (e.g., implicit, general, partial, or deflective answers).}

\textbf{Clear Non-Reply}: \textit{The respondent openly refuses to share information, claims ignorance, or requests clarification.}

Formally, let $q$ denote a question and $a$ its corresponding answer. The task is to learn a function $f_1: (q, a) \rightarrow \{1, 2, 3\}$ mapping each QA pair to one of the three clarity labels. This is a single-label, multi-class classification task.

\paragraph{Task 2 -- Evasion-level Classification:}

Given the same question--answer pair, classify the response into one of 9 fine-grained evasion techniques that form the lower level of the taxonomy. These techniques are grouped under the three clarity categories as follows: \textit{Explicit} (Clear Reply); \textit{Implicit}, \textit{General}, \textit{Partial}, \textit{Dodging}, and \textit{Deflection} (Ambivalent); and \textit{Declining to answer}, \textit{Claims ignorance}, and \textit{Clarification} (Clear Non-Reply).

The task is to learn a function $f_2: (q, a) \rightarrow \{1, \ldots, 9\}$, mapping each QA pair to one of the nine evasion labels. This is also a single-label, multi-class classification task.

Our submission addresses both tasks. Our primary leaderboard submission targets Task 1, while we additionally explore Task 2 to investigate whether fine-grained evasion classification can inform and improve clarity-level predictions.

\section{Related Work}

Recent NLP research has begun to operationalize political discourse analysis building on the theoretical frameworks of political equivocation. \citet{ferracane2021} crowdsourced annotations of political interview answers to capture subjective judgments about whether respondents intended to answer and whether their responses were truthful. In contrast, \citet{thomas2024isaidthatdataset} introduced the QEvasion dataset and a two-level taxonomy that focuses on the \textit{clarity} of responses rather than speaker intent, showing that fine-grained evasion labels can improve high-level clarity classification. Their work forms the basis of the CLARITY shared task and the dataset used in this paper.

Modeling such distinctions requires architectures capable of capturing subtle linguistic patterns while handling the inherent class imbalance in political discourse data. Transformer-based encoders such as DeBERTa-V3 \citep{he2023debertav} offer strong representational power through their disentangled attention mechanism. To address the severe imbalance between clear replies and evasive responses, we draw on focal loss \citep{lin2017focal}, which re-weights the training signal toward hard examples, and data augmentation strategies such as EDA \citep{wei2019eda} and context-aware synthetic generation \citep{park2024casa}, both of which have been shown to improve minority-class performance. We additionally employ layer-wise learning rate decay \citep{zhang2021revisitingfewsamplebertfinetuning} to preserve pretrained knowledge in lower layers while allowing task-specific adaptation in higher layers. Our system integrates these techniques to address both the linguistic complexity and data imbalance inherent in the CLARITY task.

\section{System Overview}
This section describes the system we submitted to the leaderboard, including data augmentation, model architecture, and training details.

\subsection{Data Augmentation for Class Imbalance}

The QEvasion training set \citep{thomas2024isaidthatdataset} is imbalanced: Ambivalent (59.2\%, 2,040), Clear Reply (30.5\%, 1,052), and Clear Non-Reply (10.3\%, 356). Preliminary experiments showed that this imbalance led to poor minority‑class recall (Clear Non-Reply F1 < 0.40). To mitigate it, we augmented the training data using two external large language models, Gemini 3 and Claude Sonnet 4.5, which generated synthetic examples without accessing the test set.

\textbf{Context-Aware Synthetic Generation (CASA) \citep{park2024casa}:}

Gemini 3 extracted rhetorical frames from minority classes (e.g., ``I cannot comment on ongoing diplomatic discussions'' for Clear Non-Reply) and combined them with randomized political contexts, producing 2,672 synthetic examples and perfectly balancing all classes at 2,040 each (6,120 total).

\textbf{Lexical Paraphrasing (EDA-inspired) \citep{wei2019eda}:}

Claude Sonnet 4.5 applied four operations to answer texts (synonym replacement, random insertion, random swap, random deletion, $p=0.1$), generating 1,086 synthetic examples and yielding a partially balanced distribution (Clear Reply: 1,498; Clear Non-Reply: 996; total 4,534).

We manually inspected 50 random samples to verify quality and political register. During training, we used a confidence‑weighted loss (0.5× for Claude, 0.7× for Gemini) to reduce overfitting.

\subsection{Model Selection: Why DeBERTa?}

Before settling on DeBERTa, we experimented with several transformer architectures, including DistilBERT, BERT, and Political DEBATE. While BERT provided a strong baseline (0.56 test F1), we observed that DeBERTa‑V3 and its variants consistently outperformed other models of comparable size on the development set. Its enhanced attention mechanism and improved pre-training led to a better understanding of nuanced political language. Preliminary runs with DeBERTa‑V3‑base achieved 0.64 dev F1 without any augmentation, leading us to adopt it as the foundation for our final system.

\subsection{Final Model Architecture}

Our leaderboard system is based on \texttt{microsoft/deberta-v3-base}\footnote{\url{https://huggingface.co/microsoft/deberta-v3-base}} (184M parameters). We extended it with several enhancements:

\textbf{Boolean features}: Two statistically significant binary features from the original data, \texttt{affirmative\_questions} and \texttt{multiple\_questions}, are passed through a small feature processor (Linear→ReLU→Dropout) and concatenated with the pooled transformer output before classification.

\textbf{Layer‑wise Learning Rate Decay (LLRD)}: Lower layers receive progressively smaller learning rates, scaled by $\alpha^{k}$ with $\alpha=0.9$, to preserve general knowledge while adapting higher layers.

\textbf{Focal Loss} \citep{lin2017focal}: To focus on hard examples, we use
    \begin{equation}
    FL(p_t) = -\alpha_t(1-p_t)^\gamma \log(p_t)
    \end{equation}
    where $p_t$ is the model's estimated probability for the true class, $\alpha_t$ is the inverse‑frequency class weight, and $\gamma=2.0$.
    
\textbf{Gradient accumulation}: 4 steps simulate an effective batch size of 32.

\textbf{Cosine annealing scheduler}: 15\% warmup steps followed by cosine decay.

Training runs for up to 6 epochs with early stopping (patience 3) based on validation Macro F1. Detailed hyperparameter tuning is described in Appendix~\ref{app:hyperparameters}.

\begin{table*}[t]
\centering
\caption{Processing pipeline for an example question–answer pair. The example shows how raw input is transformed before being fed to the model.}
\label{tab:example}
\begin{tabular}{@{}lp{10cm}@{}}
\toprule
\textbf{Input} & \\
\midrule
Question & Will you increase funding for education? \\
Answer   & I cannot comment on budget discussions at this time. \\
Ground truth & Clear Non-Reply \\
\midrule
\textbf{After formatting} & Question: Will you increase funding for education? [SEP] Answer: I cannot comment on budget discussions at this time. \\
\textbf{Boolean features} & \texttt{affirmative\_questions}=1, \texttt{multiple\_questions}=0 \\
\textbf{Model output} & Clear Non-Reply \\
\bottomrule
\end{tabular}
\end{table*}

\section{Experiments \&\ Results}

\subsection{Evaluation Setup}

The QEvasion dataset provides a predefined training set (3,448 samples) and test set (308 samples). We further split the training set into training (80\%, 2,758) and development (20\%, 690) using stratified sampling. Macro F1 is the official metric.

\subsection{Leaderboard System Performance}

We trained three variants of DeBERTa‑V3‑base on different data configurations (Section 4.1). Their performance on the test and evaluation datasets is shown in Table~\ref{tab:leaderboard}. The Gemini‑augmented model achieved the highest test score and was our primary submission.

For Subtask 2, our DeBERTa-V3-large model with focal loss achieved a Macro F1 of 0.45 on the development phase (rank 9 of 24 teams), but dropped sharply to 0.28 on the evaluation phase (rank 30 of 33 teams). We attribute this drop to distribution shift between phases; a detailed analysis is provided in Appendix~\ref{appendix:subtask2}.

\begin{table}[t]
\centering
\caption{Performance of DeBERTa‑V3‑base variants}
\label{tab:leaderboard}
\begin{tabular}{@{}lcc@{}}
\toprule
Configuration & Test F1 & Eval F1 \\
\midrule
Original only                & 0.64   & 0.69 \\
Claude‑augmented              & 0.65   & 0.74 \\
Gemini‑augmented (submitted)  & \textbf{0.66} & \textbf{0.76} \\
Top system (TeleAI) & -- & \textbf{0.89}\\
\bottomrule
\end{tabular}
\end{table}

For Subtask 1, our Gemini‑augmented system achieved a test Macro F1 of 0.76, placing us \textbf{8th out of 40 teams} on the leaderboard. The top system scored 0.89, while the mean score was 0.70. This puts our system in the top 20\% of participants, demonstrating the effectiveness of our augmentation and modeling choices.

\subsection{Comparison with Baselines}

Table~\ref{tab:comparison} summarizes the performance of our best system against several baselines (detailed in Appendix~\ref{app:baseline}). The Gemini‑augmented DeBERTa‑V3‑base outperforms all classical and simple transformer models. We also include a trivial baseline that always predicts the majority class (Ambivalent); its Macro F1 of 0.27 reflects the difficulty of the task and confirms that our models are learning meaningful patterns.

\subsection{Error Analysis}

To better understand the remaining errors, we examined confusion matrices for both the test set (308 samples) and the final evaluation set (237 samples) in Tables~\ref{tab:confusion_test} and~\ref{tab:confusion_eval}. The matrices reveal consistent patterns across both splits, suggesting our model generalizes well without overfitting to specific data characteristics.

\begin{table}[t]
\centering
\caption{Confusion matrix for Gemini‑augmented DeBERTa‑V3‑base on the test set (308 samples). Rows = true labels, columns = predictions. Row and column totals include percentages.}
\label{tab:confusion_test}
\resizebox{\linewidth}{!}{%
\begin{tabular}{@{}lccccc@{}}
\toprule
 & Amb & Clear & Clear-N & \textbf{R-Total} \\
\midrule
Amb       & 136 & 58  & 12 & \textbf{206} (66.9\%) \\
Clear     & 23  & 53  & 3  & \textbf{79} (25.6\%) \\
Clear-N   & 6   & 0   & 17 & \textbf{23} (7.5\%) \\
\textbf{C-Total} & \textbf{165} (53.5\%) & \textbf{111} (36.1\%) & \textbf{32} (10.4\%) & \textbf{308} \\
\bottomrule
\end{tabular}%
}
\end{table}

\begin{table}[t]
\centering
\caption{Confusion matrix for Gemini‑augmented DeBERTa‑V3‑base on the evaluation set (237 samples). Rows = true labels, columns = predictions.}
\label{tab:confusion_eval}
\resizebox{\linewidth}{!}{%
\begin{tabular}{@{}lccccc@{}}
\toprule
 & Amb & Clear & Clear-N & \textbf{R-Total} \\
\midrule
Amb       & 91 & 20  & 6 & \textbf{117} (49.4\%) \\
Clear     & 16 & 68  & 1 & \textbf{85} (35.8\%) \\
Clear-N   & 12 & 0   & 23 & \textbf{35} (14.8\%) \\
\textbf{C-Total} & \textbf{119} (50.3\%) & \textbf{88} (37.1\%) & \textbf{30} (12.6\%) & \textbf{237} \\
\bottomrule
\end{tabular}%
}
\end{table}

The test set confusion matrix shows that the main source of confusion is between Ambivalent and Clear Reply, accounting for 68\% of all errors (58 + 23 = 81 out of 119 total errors). This pattern persists in the evaluation set, where Ambivalent–Clear Reply confusion represents 65\% of errors (20 + 16 = 36 out of 55 total errors). This consistent pattern indicates that the model fundamentally struggles with distinguishing partially informative answers from fully direct ones, even after data augmentation.

The model achieves a Macro F1 of 0.6364 on the test set, with per-class F1 scores of 0.73 (Ambivalent), 0.56 (Clear Reply), and 0.62 (Clear Non-Reply). Notably, the model perfectly distinguishes Clear Non-Reply from Clear Reply in both splits (zero false positives in the Clear Reply column for true Clear Non-Reply samples), though it still misclassifies some Clear Non-Reply instances as Ambivalent.

Appendix~\ref{app:error_examples} presents multiple examples of each error type, revealing recurring patterns:

\textbf{Ambivalent → Clear Reply errors}: Answers contain both a direct statement and hedging (e.g., ``We have increased funding, but we need to study the impact further''). The model latches onto the concrete claim while missing the qualifiers.

\textbf{Clear Reply → Ambivalent errors}: Answers are direct but use tentative or conditional language (e.g., ``I believe we will consider raising funds''). The hedging vocabulary misleads the model into classifying them as Ambivalent.

\textbf{Clear Non-Reply errors}: Answers are evasive but contain factual statements or appear superficially responsive (e.g., ``The budget is complex; I cannot comment now, but here are last year's numbers''). The inclusion of concrete facts confuses the classifier.

These observations suggest that further gains could be obtained by better modeling hedging and partial answers, perhaps through multi-task learning that jointly predicts both clarity and the presence of hedging language, or by incorporating external knowledge about political discourse patterns.

\section{Additional Models}

We also experimented with two domain‑adapted or larger models; their results are summarized in Table~\ref{tab:additional_results}.

\subsection{Political DEBATE}

\texttt{mlburnham/Political\_DEBATE\_base\_v1.0}\footnote{\url{https://huggingface.co/mlburnham/Political_DEBATE_base_v1.0}} is a DeBERTa‑base model further pre‑trained on political discourse. We fine‑tuned it with the same advanced pipeline as DeBERTa‑V3‑base (LLRD, gradient accumulation, cosine annealing, early stopping), using learning rate 3e‑5 and effective batch size 32.

\subsection{DeBERTa‑V3‑large‑NLI}

\texttt{MoritzLaurer/DeBERTa-v3-large-nli}\footnote{\url{https://huggingface.co/MoritzLaurer/DeBERTa-v3-large-mnli-fever-anli-ling-wanli}} (435M parameters) is pre‑trained on multiple NLI datasets. Training followed the DeBERTa‑V3‑base pipeline but with batch size 4 and gradient accumulation steps increased to 8 (effective batch size 32). Although this model achieved comparable performance to our Gemini-augmented DeBERTa‑V3‑base, we selected the latter as our primary submission due to its lower training cost and faster iteration time.

\subsection{Additional Results}

Table~\ref{tab:additional_results} shows the performance of the additional models on the development set (test scores were not submitted).

\begin{table}[t]
\centering
\caption{Additional model results (development set)}
\label{tab:additional_results}
\begin{tabular}{lc}
\hline
Model & Test F1 \\
\hline
Political DEBATE             & 0.57 \\
DeBERTa‑V3‑large‑NLI         & 0.66 \\
\hline
\end{tabular}
\end{table}

\section{Future Work and Conclusions}

We presented an augmented DeBERTa-V3-base system for classifying the clarity of political interview responses. Our primary contribution is demonstrating that LLM-based synthetic data augmentation, combined with focal loss and careful hyperparameter tuning, can substantially improve minority-class performance on this task, raising evaluation Macro F1 from 0.69 (no augmentation) to 0.76 (Gemini-augmented). Participating in the CLARITY task reinforced that the core difficulty lies not in identifying outright non‑replies, which our model handles well, but in the gray zone between Ambivalent and Clear Reply responses, where even human annotators disagree (\(\kappa=0.65\)).

If we were to continue this work, we would pursue three directions.

First, a multi-task learning setup that jointly predicts clarity labels and evasion-level labels could leverage the hierarchical structure of the taxonomy, as prior work has shown that fine-grained evasion classification improves clarity-level predictions \citep{thomas2024isaidthatdataset}.

Second, hedging-aware features such as explicit detection of qualifiers, conditionals, and topic shifts could help the model distinguish partially informative answers from genuinely direct ones, addressing the dominant error pattern in our system.

Third, our Subtask 2 results (Appendix~\ref{appendix:subtask2}) exposed a critical need for more robust validation and minority-class augmentation at the nine-class level, where distribution shift between development and evaluation phases caused a steep performance drop.

\section{Acknowledgments}

The authors would like to thank the organizers
for the opportunity to participate in SemEval-2026
Task 6 and the Department of Computer Science,
University of Minnesota Duluth, for helping us
with all resources needed to participate in this task.
We also appreciate the valuable input and guidance
provided by the reviewers.

\bibliography{custom}

@misc{thomas2024isaidthatdataset,
  title={"I Never Said That": A dataset, taxonomy and baselines on response clarity classification},
  author={Konstantinos Thomas and Giorgos Filandrianos and Maria Lymperaiou and Chrysoula Zerva and Giorgos Stamou},
  year={2024},
  eprint={2409.13879},
  archivePrefix={arXiv},
  primaryClass={cs.CL},
  url={https://arxiv.org/abs/2409.13879},
}

@article{wei2019eda,
  title={EDA: Easy Data Augmentation Techniques for Boosting Performance on Text Classification Tasks},
  author={Wei, Jason and Zou, Kai},
  journal={arXiv preprint arXiv:1901.11196},
  year={2019},
  url={https://arxiv.org/abs/1901.11196}
}

@article{park2024casa,
  title={CASA: Context-Aware Synthetic Augmentation for Text Classification},
  author={Park, Chanwoo and others},
  journal={arXiv preprint arXiv:2403.02990},
  year={2024},
  url={https://arxiv.org/abs/2403.02990}
}

@article{lin2017focal,
  author       = {Tsung{-}Yi Lin and
                  Priya Goyal and
                  Ross B. Girshick and
                  Kaiming He and
                  Piotr Doll{\'{a}}r},
  title        = {Focal Loss for Dense Object Detection},
  journal      = {CoRR},
  volume       = {abs/1708.02002},
  year         = {2017},
  url          = {http://arxiv.org/abs/1708.02002},
  eprinttype    = {arXiv},
  eprint       = {1708.02002},
  bibsource    = {dblp computer science bibliography, https://dblp.org}
}

@article{bullequi,
author = {Bavelas, Janet and Black, Alex and Bryson, Lisa and Mullett, Jennifer},
year = {1988},
month = {06},
pages = {137-145},
title = {Political Equivocation: A Situational Explanation},
volume = {7},
journal = {Journal of Language and Social Psychology},
doi = {10.1177/0261927X8800700204}
}

@article{bullqa,
author = {Bull, Peter},
year = {1994},
month = {06},
pages = {115-131},
title = {On Identifying Questions, Replies, and Non-Replies in Political Interviews},
volume = {13},
journal = {Journal of Language and Social Psychology},
doi = {10.1177/0261927X94132002}
}

@article{bull2019cant,
author = {Bull, Peter and Strawson, Will},
year = {2019},
month = {02},
pages = {},
title = {Can’t Answer? Won’t Answer? An Analysis of Equivocal Responses by Theresa May in Prime Minister’s Questions},
volume = {73},
journal = {Parliamentary Affairs},
doi = {10.1093/pa/gsz003}
}

@misc{zhang2021revisitingfewsamplebertfinetuning,
      title={Revisiting Few-sample BERT Fine-tuning}, 
      author={Tianyi Zhang and Felix Wu and Arzoo Katiyar and Kilian Q. Weinberger and Yoav Artzi},
      year={2021},
      eprint={2006.05987},
      archivePrefix={arXiv},
      primaryClass={cs.CL},
      url={https://arxiv.org/abs/2006.05987}, 
}

@inproceedings{
he2023debertav,
title={De{BERT}aV3: Improving De{BERT}a using {ELECTRA}-Style Pre-Training with Gradient-Disentangled Embedding Sharing},
author={Pengcheng He and Jianfeng Gao and Weizhu Chen},
booktitle={The Eleventh International Conference on Learning Representations },
year={2023},
url={https://openreview.net/forum?id=sE7-XhLxHA}
}

@inproceedings{ferracane2021,
    title = "Did they answer? Subjective acts and intents in conversational discourse",
    author = "Ferracane, Elisa  and
      Durrett, Greg  and
      Li, Junyi Jessy  and
      Erk, Katrin",
    editor = "Toutanova, Kristina  and
      Rumshisky, Anna  and
      Zettlemoyer, Luke  and
      Hakkani-Tur, Dilek  and
      Beltagy, Iz  and
      Bethard, Steven  and
      Cotterell, Ryan  and
      Chakraborty, Tanmoy  and
      Zhou, Yichao",
    booktitle = "Proceedings of the 2021 Conference of the North American Chapter of the Association for Computational Linguistics: Human Language Technologies",
    month = jun,
    year = "2021",
    address = "Online",
    publisher = "Association for Computational Linguistics",
    url = "https://aclanthology.org/2021.naacl-main.129/",
    doi = "10.18653/v1/2021.naacl-main.129",
    pages = "1626--1644"
}

\appendix
\section{Hyperparameter Tuning Details}
\label{app:hyperparameters}

We performed a focused hyperparameter search on the original training set, fixing gradient accumulation at 4 and $\gamma=2.0$. We varied the learning rate and LLRD factor:
\begin{itemize}
    \item Learning rates tested: 2e‑5, 3e‑5, 5e‑5 (with $\alpha=0.9$). The best was 3e‑5.
    \item With learning rate 3e‑5, we explored $\alpha=0.8$, $0.9$, $0.95$. $\alpha=0.9$ gave the most stable training.
\end{itemize}
These values were used for all subsequent experiments, including the augmented‑data variants.

\section{Baseline Models}
\label{app:baseline}
Here we describe the configurations of models that served as baselines.

\subsection{Classical Machine Learning Models}

\textbf{TF‑IDF + Logistic Regression} – scikit‑learn’s \texttt{TfidfVectorizer} with \texttt{max\_features=5000}, \texttt{ngram\_range=(1,2)}, \texttt{min\_df=2}, \texttt{max\_df=0.95}, \texttt{stop\_words='english'}. Classifier: LogisticRegression(\texttt{class\_weight='balanced'}, \texttt{C=1.0}).

\textbf{SVM} – Same TF‑IDF with L2 normalization and sublinear tf scaling. Grid search over \texttt{C=[0.1,1.0,10.0]}, kernels (\texttt{linear},\texttt{rbf}), 3‑fold CV, \texttt{class\_weight='balanced'}.

\textbf{Random Forest} – 100 trees, \texttt{max\_depth=20}, \texttt{min\_samples\_split=10}, \texttt{min\_samples\_leaf=4}, \texttt{class\_weight='balanced'}.

\subsection{Simple Transformers}

All transformers were fine‑tuned with HuggingFace Transformers using AdamW (weight decay 0.01), gradient clipping (1.0), linear warmup (10\%) and linear decay, and class‑weighted cross‑entropy.

\textbf{DistilBERT} – \texttt{distilbert-base-uncased}\footnote{\url{https://huggingface.co/distilbert-base-uncased}}, 4 epochs, batch size 16, learning rate 2e‑5.

\textbf{BERT} – \texttt{bert-base-uncased}\footnote{\url{https://huggingface.co/bert-base-uncased}}, 4 epochs, batch size 8 (due to memory), learning rate 2e‑5.

\begin{table}[t]
\centering
\caption{Comparison with baseline models}
\label{tab:comparison}
\begin{tabular}{@{}lcc@{}}
\toprule
Model & Test F1 \\
\midrule
Majority class (Ambivalent)   & 0.2700  \\
TF‑IDF + Logistic Regression   & 0.4476 \\
SVM (linear)                   & 0.4270 \\
Random Forest                   & 0.4256 \\
DistilBERT                      & 0.5158 \\
BERT‑base                       & 0.5628 \\
\midrule
\textbf{DeBERTa‑V3‑base (Gemini)} & \textbf{0.66}\\
\bottomrule
\end{tabular}
\end{table}

\section{Detailed Error Examples}
\label{app:error_examples}

This appendix provides three representative examples for each of the main error types observed in the test set predictions of our Gemini‑augmented DeBERTa‑V3‑base model. The examples illustrate the recurring patterns that cause misclassifications.

\subsection{Ambivalent → Clear Reply Errors (20 total)}

The model over‑commits to a label when the response sounds direct but does not fully address the question.

\begin{itemize}
    \item \textbf{Example 1 (June 2017):} \\
    \textit{Question:} ``So he said those things under oath?'' \\
    \textit{Answer:} ``One-hundred percent. I didn't say under oath—I hardly know the man... No, I didn't say that, and I didn't say the other.'' \\
    \textit{Analysis:} The opening ``One-hundred percent'' signals directness, but the answer neither confirms nor denies the claim—it pivots to what the speaker did not say. The model latches onto the confident opening and ignores the subsequent hedging.

    \item \textbf{Example 2 (February 2016):} \\
    \textit{Question:} ``Would you consider a recess appointment if your nominee is not granted a hearing?'' \\
    \textit{Answer:} ``I think that we have more than enough time to go through regular order... I expect them to hold hearings. I expect there to be a vote.'' \\
    \textit{Analysis:} The response never actually answers the yes/no question about a recess appointment. The model is fooled by the confident declarative tone and the detailed discussion of the regular process.

    \item \textbf{Example 3 (June 2006):} \\
    \textit{Question:} ``Do you have a specific target for how much you want the violence reduced?'' \\
    \textit{Answer:} ``Enough for the Government to succeed.'' \\
    \textit{Analysis:} The answer sounds direct but is entirely non‑specific. The question asks for a numerical target, and the response gives a vague criterion. The model mistakes this for a Clear Reply because of its assertive phrasing.
\end{itemize}

\subsection{Clear Reply → Ambivalent Errors (16 total)}

The model under‑commits when responses are brief, blunt, or terse—mistaking conciseness for hedging.

\begin{itemize}
    \item \textbf{Example 1 (January 1953):} \\
    \textit{Question:} ``You have described it as a billion dollar steal?'' \\
    \textit{Answer:} ``You left off two zeros. It's a hundred billion dollars.'' \\
    \textit{Analysis:} This is a crisp, direct correction. The model likely penalizes the brevity or the indirectness of the corrective framing (a question–answer that corrects without a simple ``yes'').

    \item \textbf{Example 2 (September 1980):} \\
    \textit{Question:} ``Does an apology rule out the question of honor?'' \\
    \textit{Answer:} ``Yes. The United States is not going to apologize.'' \\
    \textit{Analysis:} An unambiguous ``Yes'' followed by a clear statement. The longer elaboration after the ``Yes'' may have confused the model into interpreting the response as more complex than a simple affirmation.

    \item \textbf{Example 3 (July 2022):} \\
    \textit{Question:} ``So you don't expect to bring up human rights?'' \\
    \textit{Answer:} ``I will bring up—I always bring up human rights. I always bring up human rights.'' \\
    \textit{Analysis:} A direct contradiction of the premise. The repetition and the interjection from another speaker (visible in the transcript) may have introduced spurious signals of ambiguity, causing the model to miss the clear reply.
\end{itemize}

\subsection{Clear Non‑Reply → Ambivalent Errors (12 total)}

The model fails to recognize explicit refusals or deflections as non‑replies, possibly because the responses contain some substantive‑sounding content or are too short.

\begin{itemize}
    \item \textbf{Example 1 (July 2017):} \\
    \textit{Question:} ``And about his role in Syria and the region?'' \\
    \textit{Answer:} ``Whose role?'' \\
    \textit{Analysis:} A pure deflection—answering a question with a question. Very short responses may not provide enough signal for the model to confidently classify them as non‑replies.

    \item \textbf{Example 2 (August 1992):} \\
    \textit{Question:} ``But they were ready to move sooner if asked, weren't they?'' \\
    \textit{Answer:} ``I'm not going to go into that because... what you seem to be interested in is kind of assigning blame. That is not what's at stake here, and I don't want to participate in that.'' \\
    \textit{Analysis:} An explicit refusal to answer—but the model may have read the surrounding substantive content (the discussion of blame) as partial engagement, leading to an Ambivalent prediction.

    \item \textbf{Example 3 (January 1953):} \\
    \textit{Question:} ``Mr. President, there is one question that is left unanswered.'' \\
    \textit{Answer:} ``What's that?'' \\
    \textit{Analysis:} Another question‑as‑response, deflecting entirely. The extreme brevity leaves the model without enough context to detect the non‑reply pattern.
\end{itemize}

\section{Subtask 2 Evaluation Phase Analysis}
\label{appendix:subtask2}
Our Subtask 2 system exhibited a substantial performance gap between the development phase (Macro F1 = 0.45, rank 9/24) and the evaluation phase (Macro F1 = 0.28, rank 30/33). Post-hoc analysis identified the following:

We identified a flaw in validation during the development phase: the original training script assigned dummy labels (all zeros) to the held-out test split, meaning early stopping and model selection were optimized against meaningless validation metrics. The model checkpoint saved as ``best'' was effectively selected at a random epoch. Despite retraining with a proper stratified cross-validation split, which raised internal validation F1 to 0.51, the evaluation phase score remained at 0.28, suggesting substantial distribution shift between the development and evaluation sets. The nine-class imbalance (13.3$\times$ ratio between the largest and smallest class) likely exacerbated this shift, as minority classes (\textit{Partial/half-answer}: 79 training samples, \textit{Clarification}: 92) are highly sensitive to domain variation. Future work should address this through cross-domain validation and more aggressive minority-class augmentation.



\end{document}